\pdfoutput=1

\documentclass[11pt]{article}

\usepackage[]{acl}

\usepackage{times}
\usepackage{latexsym}

\usepackage[T1]{fontenc}

\usepackage[utf8]{inputenc}

\usepackage{microtype}

\usepackage{enumitem}
\usepackage{bookmark}
\usepackage{bm}
\usepackage{amsfonts}
\usepackage{subfigure}
\usepackage{graphicx}
\usepackage{booktabs}
\usepackage{makecell}
\usepackage{xcolor}
\usepackage{amsmath}
\usepackage{arydshln}
\usepackage{multirow}
\usepackage{stmaryrd}
\usepackage{color}
\usepackage{makecell}

\newcommand{\rmark}[1]{\textcolor{red}{#1}}

\newcommand{\tran}{\top}
\newcommand\blfootnote[1]{%
  \begingroup
  \renewcommand\thefootnote{}\footnote{#1}%
  \addtocounter{footnote}{-1}%
  \endgroup
}

\usepackage{bm}
\title{Improving Event Representation via Simultaneous Weakly Supervised Contrastive Learning  and Clustering }

\author{ Jun Gao\textsuperscript{1}\enskip Wei Wang\textsuperscript{3}\enskip Changlong Yu\textsuperscript{4}\enskip Huan Zhao\textsuperscript{5}\enskip Wilfred Ng\textsuperscript{4}\enskip Ruifeng Xu\textsuperscript{1,2}\thanks{\;\;Corresponding author}\\
\normalsize \textsuperscript{1}Harbin Institute of Technology (Shenzhen)\quad \textsuperscript{2}Peng Cheng Laboratory\quad \textsuperscript{3}Tsinghua University \\
\normalsize \texttt{imgaojun@gmail.com}\quad \texttt{xuruifeng@hit.edu.cn}\quad \texttt{weiwangorg@163.com}\\
\normalsize\quad \textsuperscript{4}HKUST, Hong Kong, China\quad \textsuperscript{5}4Paradigm. Inc. \\
\normalsize \texttt{\{cyuaq,wilfred\}@cse.ust.hk}\quad \texttt{zhaohuan@4paradigm.com}\\
}

\begin{document}
\maketitle
\begin{abstract}

Representations of events described in text are important for various tasks. In this work, we present \textbf{SWCC}: a \textbf{S}imultaneous \textbf{W}eakly supervised \textbf{C}ontrastive learning and \textbf{C}lustering framework for event representation learning. SWCC learns event representations by making better use of co-occurrence information of events. Specifically, we introduce a weakly supervised contrastive learning method that allows us to consider multiple positives and multiple negatives, and a prototype-based clustering method that avoids semantically related events being pulled apart. For model training, SWCC learns representations by simultaneously performing weakly supervised contrastive learning and prototype-based clustering.
Experimental results show that SWCC outperforms other baselines on \texttt{Hard Similarity} and \texttt{Transitive Sentence Similarity} tasks. In addition, a thorough analysis of the prototype-based clustering method demonstrates that the learned prototype vectors are able to implicitly capture various relations between events.
Our code will be available at \url{https://github.com/gaojun4ever/SWCC4Event}.
\blfootnote{Jun Gao is currently a research intern at 4Paradigm. }
\end{abstract}

\section{Introduction}
\begin{figure}[htbp] 
  \centering
  \includegraphics[width=1.0\linewidth]{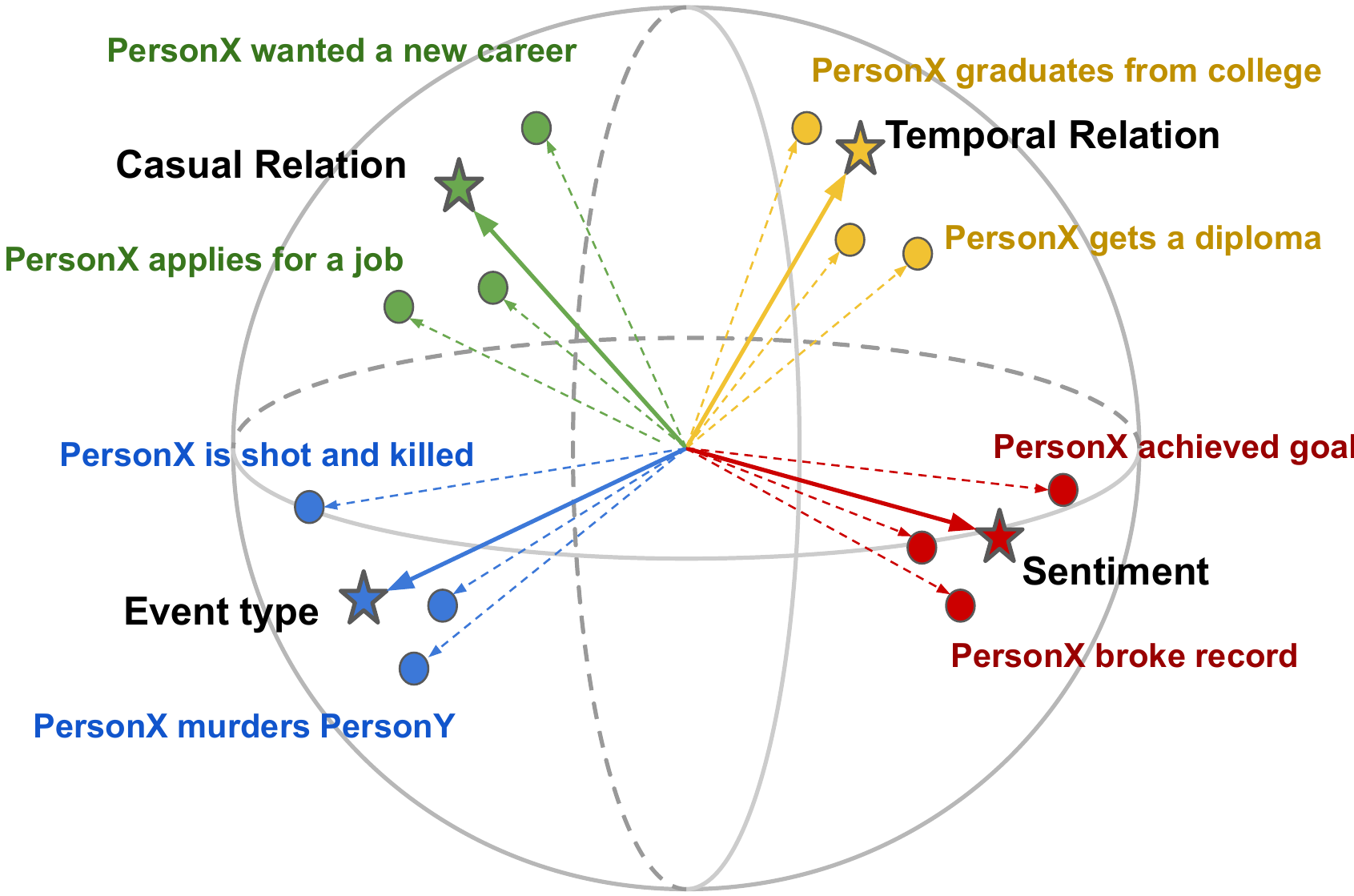}
  \caption{Four pairs of events with different relations. Stars represent prototypes and circles represent events.}
  \vspace{-1.5em}
  \label{fig:example}
\end{figure}

Distributed representations of events, are a common way to represent events in a machine-readable form and have shown to provide meaningful features for various tasks~\citep{lee2018feel,rezaee2020event,deng2021ontoed,martin2018event,chen2021graphplan}.
Obtaining effective event representations is challenging, as it requires representations to capture various relations between events. 
Figure~\ref{fig:example} presents four pairs of events with different relations. 
Two events may share the same event attributes (e.g. event types and sentiments), and there may also be a causal or temporal relation between two events.

Early works~\citep{weber2018event} exploit easily accessible co-occurrence relation of events to learn event representations. Although the use of co-occurrence relation works well, it is too coarse for deep understanding of events, which requires fine-grained knowledge~\citep{lee-goldwasser-2019-multi}.
Recent works focus on fine-grained knowledge, such as discourse relations~\citep{lee-goldwasser-2019-multi,zheng2020incorporating} and commonsense knowledge~(e.g. sentiments and intents)~\citep{sap2019atomic,ding-etal-2019-event-representation}. Concretely, \citet{lee-goldwasser-2019-multi} and \citet{zheng2020incorporating} leverage 11 discourse relation types to model event script knowledge. \citet{ding-etal-2019-event-representation} incorporate manually labeled commonsense knowledge~(intents and sentiments) into event representation learning.
However, the types of fine-grained event knowledge are so diverse that we cannot enumerate all of them and currently adopted fine-grained knowledge fall under a small set of event knowledge.
In addition, some manually labeled knowledge~\citep{sap2019atomic,hwang2021symbolic} is costly and difficult to apply on large datasets.

In our work, we observe that there is a rich amount of information in co-occurring events, but previous works did not make good use of such information. Based on existing works on event relation extraction~\citep{Xue2016CoNLL2S,lee-goldwasser-2019-multi,zhang2020aser,Wang2020JointCL}, we find that the co-occurrence relation, which refers to two events appearing in the same document,  can be seen as a superset of currently defined explicit discourse relations. To be specific, these relations are often indicated by discourse markers (e.g., ``because'', capturing the 
casual relation)~\citep{lee-goldwasser-2019-multi}. Therefore, two related events must exist in the same sentence or document. More than that, the co-occurrence relation also includes other implicit event knowledge. For example, events that occur in the same document may share the same topic and event type. To learn event representations, previous works~\citep{granroth2016happens,weber2018event} based on co-occurrence information usually exploit instance-wise contrastive learning approaches related to the margin loss, which consists of an anchor, positive, and negative sample, where the anchor is more similar to the positive than the negative.
However, they share two common limitations: 
 (1)~such margin-based approaches struggle to capture the essential differences between events with different semantics, as they only consider one positive and one negative per anchor. 
 (2)~Randomly sampled negative samples may contain samples semantically related to the anchor, but are undesirably pushed apart in embedding space. This problem arises because these instance-wise contrastive learning approaches treat randomly selected events as negative samples, regardless of their semantic relevance.


We are motivated to address the above issues with the goal of  making better use of co-occurrence information of events. To this end, we present \textbf{SWCC}: 
a \textbf{S}imultaneous \textbf{W}eakly supervised \textbf{C}ontrastive learning and \textbf{C}lustering framework for event representation learning, where we exploit document-level co-occurrence information of events as weak supervision and learn event representations by simultaneously performing weakly supervised contrastive learning and prototype-based clustering. 
To address the first issue, we build our approach on  the contrastive framework with the InfoNCE  objective~\citep{oord2019representation}, which is a self-supervised contrastive learning method that uses one positive and multiple negatives. Further, we extend the InfoNCE to a weakly supervised contrastive learning setting, allowing us to consider multiple positives and multiple negatives per anchor~(as opposed to the previous works which use only one positive and one negative). Co-occurring events are then incorporated as additional positives, weighted by a normalized co-occurrence frequency. To address the second issue, we introduce a prototype-based clustering method to avoid semantically related events being pulled apart. Specifically, we impose a prototype for each cluster, which is a representative embedding for a group of semantically related events. Then we cluster the data while enforce consistency between cluster assignments produced for different augmented representations of an event.
Unlike the instance-wise contrastive learning, our clustering method focuses on the cluster-level semantic concepts by contrasting between representations of events and clusters. 
Overall, we make the following contributions:
\begin{itemize}[leftmargin=*,noitemsep,nolistsep]
    \item We propose a simple and effective framework~(\textbf{SWCC}) that learns event representations by making better use of co-occurrence information of events. Experimental results show that our approach outperforms previous approaches on several event related tasks.
    \item We introduce a weakly supervised contrastive learning method that allows us to consider multiple positives and multiple negatives, and a prototype-based clustering method that avoids semantically related events being pulled apart.
    \item  We provide a thorough analysis of the prototype-based clustering method to demonstrate that the learned prototype vectors are able to implicitly capture various relations between events.
\end{itemize}

\begin{figure*}[htbp] 
    \centering
    \includegraphics[width=0.8\linewidth]{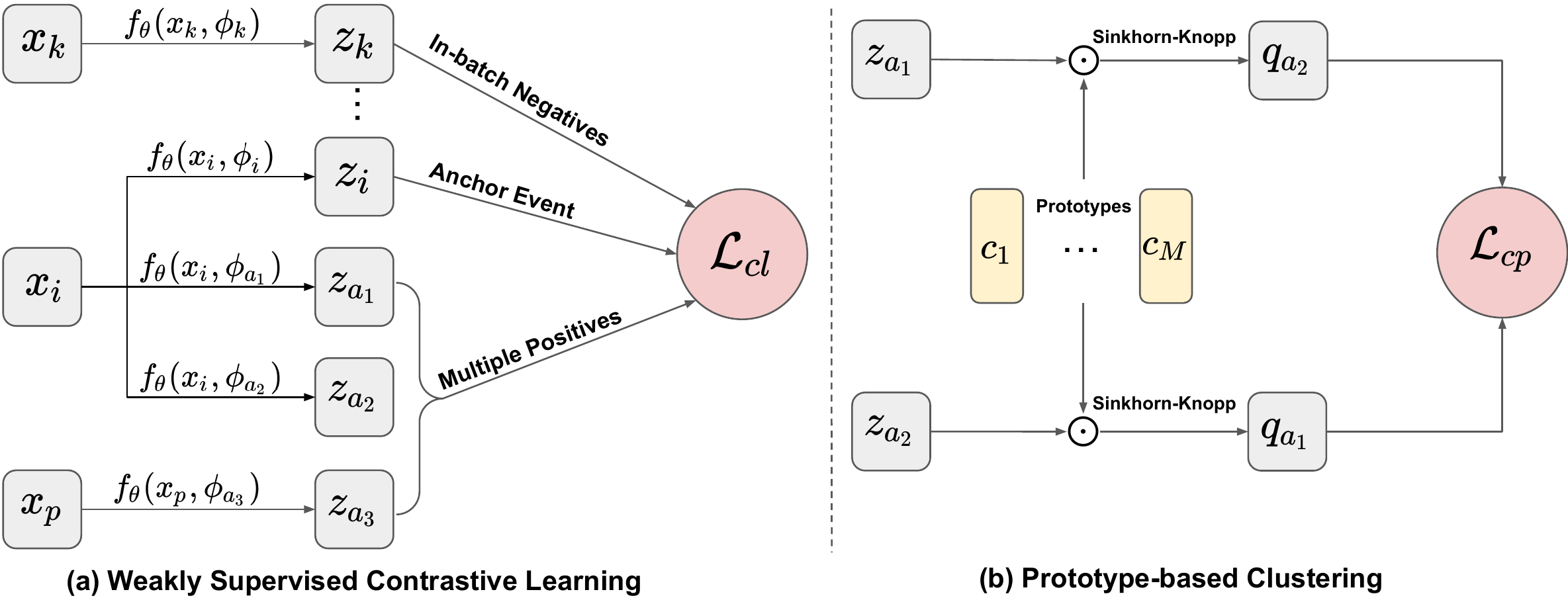}
    \caption{Architecture of the proposed framework, where the left part is the \texttt{Weakly Supervised Contrastive Learning} method and the right part is the \texttt{Prototype-based Clustering} method. Given an input event $\bm{x}_i$, we obtain three augmented representations $\bm{z}_i,\bm{z}_{a_1}$ and $\bm{z}_{a_2}$ of the same event $\bm{x}_i$ using the BERT model with different dropout masks. Using the same approach, we obtain the representation set $\{\bm{z}_k\}_{k\in\mathcal{N}(i)}$ of in-batch negatives and the representation $\bm{z}_{a_3}$ of its co-occurrence event.}
    \label{fig:arch}
    \vspace{-1em}
  \end{figure*}

\section{Preliminaries}
\paragraph{Event representation model.}
\label{sec:model}
In the early works~\citep{weber2018event,ding-etal-2019-event-representation}, Neural Tensor Networks (NTNs)~\citep{socher2013recursive,socher2013reasoning} are widely adopted to compose the representation of event constitutions, i.e., \texttt{(subject, predicate, object)}. 
However, such methods introduced strong compositional inductive bias and can not extend to events with more additional arguments, such as time, location etc. 
Several recent works~\citep{zheng2020incorporating,vijayaraghavan-roy-2021-lifelong} replaced static word vector compositions with  powerful pretrained language models, such as BERT~\cite{devlin-etal-2019-bert}, for flexible event representations and achieved better performance. 
Following them, we also take the BERT as the backbone model. 

The BERT encoder can take as input a free-form event text, which contains a sequence of tokens and the input format can be represented as follows:
\begin{equation}
    [\mathrm{CLS}], pred, subj, obj, [\mathrm{SEP}].
\end{equation}
Define $\bm{x} = [x_0,x_1,\cdots,x_L]$ to be the input sequence of length $L$, where $x_0$ and $x_L$ are the [CLS] token and the [SEP] token respectively. Given $\bm{x}$, the BERT returns a sequence of contextualized vectors:
\begin{equation}
    [\bm{v}_{[\mathrm{CLS}]}, \bm{v}_{x_1}, \cdots,\bm{v}_{x_L}] = \mathrm{BERT}(\bm{x}),
\end{equation}
where $\bm{v}_{[\mathrm{CLS}]}$ is the representation for the [CLS] token. In the default case, the final vector representation $\bm{z}$ of the event is the output representation of the [CLS] token: $\bm{z} = \bm{v}_{[\mathrm{CLS}]}$.

\paragraph{Instance-wise contrastive learning.}
Event representation models learn representations with contrastive learning, which aims to pull related events together and push apart unrelated events. 
Margin loss~\citep{schroff2015facenet} is a widely used contrastive loss in most of the existing works on event representation learning~\citep{weber2018event,ding-etal-2019-event-representation,zheng2020incorporating}. Most recently, an alternative contrastive loss function, called InfoNCE~\citep{oord2019representation}, has been proposed and shown effective in various contrastive learning tasks~\citep{he2020momentum,hu2021adco,gao2021simcse}. \citet{chen2020simple} further demonstrate that InfoNCE works better than the Margin loss.
In this work, we explore the use of InfoNCE to train our event representation model. 

Formally, given a set of $N$ paired events $\mathcal{D}=\{\bm{x}_i,\bm{x}_i^+\}_{i=1}^N$, where $\bm{x}_i^+$ is a positive sample for $\bm{x}_i$, the InfoNCE objective for $(\bm{x}_i,\bm{x}_i^+)$ is presented in a softmax form with in-batch negatives~\citep{chen2020simple,gao2021simcse}:
\begin{equation}
    \mathcal{L}= -\mathrm{log}\frac{g(\bm{z}_i,\bm{z}_i^+)}{g(\bm{z}_i,\bm{z}_i^+)+\sum_{k\in \mathcal{N}(i)} g(\bm{z}_i,\bm{z}_k)},
    \label{eq:infonce}
\end{equation}
where $\bm{z}_i$ and $\bm{z}_i^+$ are the augmented representations of $\bm{x}_i$ and $\bm{x}_i^+$ obtained through a representation model
, $k\in \mathcal{N}(i)$ is the index of in-batch negatives.
and $g$ is a function: $g(\bm{z}_i,\bm{z}_k)=\exp(\bm{z}_i^\tran\bm{z}_k/\tau)$, where $\tau \in \mathbb{R}^+$ is a positive value of temperature.

\paragraph{Data augmentation.} 
One critical question in contrastive learning is how to obtain $\bm{z}_i^+$. In language representation, $\bm{z}_i^+$ are often obtained by first applying data augmentation in the form of word deletion, reordering, or substitution on $\bm{x}_i$ and then feeding it into the event representation model. Several recent works~\citep{gao2021simcse,liang2021r} exploit dropout noise as data augmentation for NLP tasks and find that this data augmentation technique performs much better than common data augmentation techniques. Specifically, given an input event $\bm{x}_i$, we obtain $\bm{z}_i$ and $\bm{z}_i^+$ by feeding the same input to the BERT encoder with the parametric weights $\theta$ twice, and each time we apply a different dropout mask:
\begin{equation}
    \bm{z}_i = f_{\theta}(\bm{x}_i,\bm{\phi}_1), \bm{z}_i^+ = f_{\theta}(\bm{x}_i,\bm{\phi}_2),
\end{equation}
where  $\bm{\phi}_1$ and $\bm{\phi}_2$ are two different random masks for dropout.
As described in Sec.\ref{sec:wscl}, given an anchor event $\bm{z}_i$ , we generate 3 positive samples $\bm{z}_{a_1}$, $\bm{z}_{a_2}$ and $\bm{z}_{a_3}$ with different dropout masks.

\section{The Proposed Approach}

In this section, we will present technical details of our proposed approach and our goal is to learn event representations by making better use of co-occurrence information of events.
Figure~\ref{fig:arch} presents an overview of our proposed approach, which contains two parts: the weakly-supervised contrastive learning method~(left) and the prototype-based clustering method~(right). 
In the following sections, we will introduce both methods separately.

\subsection{Weakly Supervised Contrastive Learning}
\label{sec:wscl}
We build our approach on the contrastive framework with the InfoNCE  objective~(Eq.\ref{eq:infonce}) instead of the margin loss.
To incorporate co-occurrence information into event representation learning, a straightforward way is to consider the co-occurring event of each input event as an additional positive sample, that is, the positive augmented representations of $\bm{x}_i$ come not only from itself but also from its co-occurring event denoted as $\bm{x}_p$.
However, The original InfoNCE objective  cannot handle the case where there exists multiple positive samples. Inspired by \citet{khosla2020supervised}, we take a similar formulation to tackle this problem. More than that, we also introduce a weighting mechanism to consider co-occurrence frequency of two events, which indicates the strength of the connection between two events. 

\paragraph{Co-occurrence as weak supervision.}
Formally, for each input pair $(\bm{x}_i,\bm{x}_p)$, where $\bm{x}_i$ and $\bm{x}_p$ refer to the input event and one of its co-occurring events, we first compute an augmented representation $\bm{z}_i$ of $\bm{x}_i$ as an anchor event, through the event representation model mentioned in \S~\ref{sec:model}. How the method differs from InfoNCE is in the construction of the positive set $\mathcal{A}(i)$ for $\bm{x}_i$. In InfoNCE, $\mathcal{A}(i)$ only contains one positive.
In our method, we generalize Eq.~\ref{eq:infonce} to support multiple positives learning:
\begin{equation}
    \mathcal{L} = \!\!\sum_{a\in \mathcal{A}(i) }\!\!-\mathrm{log}\frac{g(\bm{z}_i,\bm{z}_a)}{g(\bm{z}_i,\bm{z}_a)+\sum_{k\in \mathcal{N}(i)} g(\bm{z}_i,\bm{z}_k)},\!\!
    \label{eq:mutipos}
\end{equation}
where $\mathcal{A}(i)$ and $\mathcal{N}(i)$ refer to the positive set  and the negative set for the event $\bm{x}_i$. Note that we support arbitrary number of positives here. In our work, considering the limited GPU memory, we use $\mathcal{A}(i)=\{\bm{z}_{a_1},\bm{z}_{a_2},\bm{z}_{a_3}\}$, where $\bm{z}_{a_1}$ and $\bm{z}_{a_2}$ are two augmented representations of the same event $\bm{x}_i$, obtained with different dropout masks, and $\bm{z}_{a_3}$ is an augmented representation of its co-occurring event. Here $\bm{z}_{a_1}$ and $\bm{z}_{a_2}$ will then be used in the prototype-based clustering method~(See Fig.~\ref{fig:arch} for example) as detailed later~(\S~\ref{sec:self_train}).

\paragraph{Incorporating co-occurrence frequency.}
The co-occurrence frequency indicates the strength of the connection between two events. To make better use of data, we introduce a weighting mechanism to exploit the co-occurrence frequency between events as instance weights and rewrite the Eq.~\ref{eq:mutipos}:
\begin{equation}
    \!\mathcal{L}_{cl} = \!\!\!\!\sum_{a\in \mathcal{A}(i) }\!\!\!\!-\mathrm{log}\frac{\varepsilon_a \cdot g(\bm{z}_i,\bm{z}_a)}{g(\bm{z}_i,\bm{z}_a)+\sum_{k\in \mathcal{N}(i)} g(\bm{z}_i,\bm{z}_k)}.\!\!\!\!
\end{equation}
Here $\varepsilon_a$ is a weight for the positive sample $\bm{z}_a$. In our work, the two weights $\varepsilon_{a_1}$ and $\varepsilon_{a_2}$ of the positive samples~($\bm{z}_{a_1}$ and $\bm{z}_{a_2}$) obtained from the input event, are set as $\varepsilon_{a_1}=\varepsilon_{a_2}=\frac{1}{|\mathcal{A}(i)|-1}$, where $|\mathcal{A}(i)|$ is its cardinality.
To obtain the weight $\varepsilon_{a_3}$ for the augmented representation $\bm{z}_{a_3}$ of the co-occurring event, we create a co–occurrence matrix, $\bm{V}$ with each entry corresponding to the co-occurrence frequency of two distinct events. Then $\bm{V}$ is normalized to $\hat{\bm{V}}$ with the \texttt{Min-Max} normalization method, and we take the entry in $\hat{\bm{V}}$ as the weight $\varepsilon_{a_3}$ for the co-occurrence event. In this way, the model draws the input events closer to the events with higher co-occurrence frequency, as each entry in $\hat{\bm{V}}$ indicates the strength of the connection between two events.

\subsection{Prototype-based Clustering}
\label{sec:self_train}

To avoid semantically related events being pulled apart, we draw inspiration from the recent approach~\citep{caron2020unsupervised} in the computer vision domain and introduce a prototype-based clustering method, where we impose a prototype, which is a representative embedding for a group of semantically related events for each cluster. Then we cluster the data while enforce consistency between cluster assignments produced for different augmented representations of an event. These prototypes
essentially serve as the center of data representation clusters for a group of semantically related events~(See Figure~\ref{fig:example} for example).
Unlike the instance-wise contrastive learning, our clustering method focuses on the cluster-level semantic concepts by contrasting between representations of events and clusters. 

  \paragraph{Cluster prediction.}
  This method works by comparing two different augmented representations of the same event using their intermediate cluster assignments. The motivation is that if these two representations capture the same information, it should be possible to predict the cluster assignment of one augmented representation from another augmented representation. In detail, we consider a set of $M$ prototypes, each associated with a learnable vector $\bm{c}_i$, where $i\in \llbracket M \rrbracket$. Given an input event, we first transform the event into two augmented representations with two different dropout masks. Here we use the two augmented representations $\bm{z}_{a_1}$ and $\bm{z}_{a_2}$ of the event $\bm{x}_i$. We compute their cluster assignments $\bm{q}_{a_1}$ and $\bm{q}_{a_2}$ by matching the two augmented representations to the set of $M$ prototypes. The cluster assignments are then swapped between the two augmented representations: the cluster assignment $\bm{q}_{a_1}$ of the augmented representation $\bm{z}_{a_1}$ should be predicted from the augmented representation $\bm{z}_{a_2}$, and vice-versa. Formally, the cluster prediction loss is defined as: 
\begin{equation}
    \mathcal{L}_{cp} = \ell(\bm{z}_{a_1},\bm{q}_{a_2}) + \ell(\bm{z}_{a_2},\bm{q}_{a_1}),
\end{equation}
where function $\ell(\bm{z},\bm{q})$ measures the fit between the representation $\bm{z}$ and the cluster assignment $\bm{q}$, as defined by: $\ell(\bm{z},\bm{q}) = -\bm{q}\mathrm{log}\bm{p}$.
Here
$\bm{p}$ is a probability vector over the $M$ prototypes whose components are:
\begin{equation}
    p^{(j)} = \frac{\exp(\bm{z}^\tran \bm{c}_j/\tau)}{\sum_{k=1}^M\exp(\exp(\bm{z}^\tran \bm{c}_k/\tau)},
\end{equation}
where $\tau$ is a temperature hyperparameter.
Intuitively, this cluster prediction method links representations $\bm{z}_{a_1}$ and $\bm{z}_{a_2}$ using the intermediate cluster assignments $\bm{q}_{a_1}$ and $\bm{q}_{a_2}$.

\paragraph{Computing cluster assignments.}
We compute the cluster assignments using an Optimal Transport solver. This solver ensures equal partitioning of the prototypes or clusters across all augmented representations, avoiding trivial solutions where all representations are mapped to a unique prototype. In particular, we employ the Sinkhorn-Knopp algorithm~\citep{cuturi2013sinkhorn}. 
The algorithm first begins with a matrix $\bm{\Gamma}  \in \mathbb{R}^{M\times N}$ with each element initialized to $\bm{z}_{b}^\tran \bm{c}_m$, where $b \in \llbracket N \rrbracket$ is the index of each column.
It then iteratively produces a doubly-normalized matrix, the columns of which comprise $\bm{q}$ for the minibatch.

\begin{table*}[htbp]
\small
    \centering
        \begin{tabular}{lrrr}
        \toprule
        \multirow{ 2}{*}{Model} & \multicolumn{2}{c}{Hard similarity (Accuracy \%)} & Transitive sentence\\\cline{2-3}
        &  Original & Extended & similarity ($\rho$)\\
        \midrule
        Event-comp~\citep{weber2018event}* & 33.9 & 18.7 & 0.57\\
        Predicate Tensor~\citep{weber2018event}* & 41.0 & 25.6 & 0.63\\
        Role-factor Tensor~\citep{weber2018event}* & 43.5 & 20.7 & 0.64\\
        \hdashline
        KGEB~\citep{ding2016knowledge}* & 52.6 & 49.8 & 0.61\\
        NTN-IntSent~\citep{ding-etal-2019-event-representation}* & 77.4 & 62.8 & 0.74\\
        \hdashline
        SAM-Net~\citep{lv2019sam}* & 51.3 & 45.2 & 0.59\\
        FEEL~\citep{lee2018feel}* & 58.7 & 50.7 & 0.67\\
        UniFA-S~\citep{zheng2020incorporating}* & 78.3 &64.1 & 0.75\\
        \hdashline
        SWCC& \textbf{80.9} & \textbf{72.1} & \textbf{0.82}\\
        \bottomrule
        \end{tabular}
        \caption{Evaluation performance on the similarity tasks. Best results are bold. *: results reported in the original papers.}
        \vspace{-1em}
        \label{tab:similarity}
\end{table*}

\subsection{Model Training}
Our approach learns event representations by simultaneously performing weakly supervised contrastive learning and prototype-based clustering. The overall training objective has three terms:
\begin{equation}
    \mathcal{L}_{overall} = \mathcal{L}_{cl} + \beta \mathcal{L}_{cp} + \gamma \mathcal{L}_{mlm},
    \label{eq:overal_loss}
\end{equation}
where $\beta$ and $\gamma$ are hyperparameters. The first term is the weakly supervised contrastive learning loss that allows us to effectively incorporate co-occurrence information into event representation learning. The second term is the prototype-based clustering loss, whose goal is to cluster the events while enforcing consistency between cluster assignments produced for different augmented representations of the input event. 
Lastly, we introduce the masked language modeling~(MLM) objective~\citep{devlin-etal-2019-bert} as an auxiliary loss to avoid forgetting of token-level knowledge. 

\section{Experiments}
Following common practice in event representation learning~\citep{weber2018event,ding-etal-2019-event-representation,zheng2020incorporating}, we analyze the event representations learned by our approach on two event similarity tasks~(\S~\ref{sec:similarity_task}) and one transfer task~(\S~\ref{sec:transfer_task}).

\subsection{Dataset and Implementation Details}
The event triples we use for the training data are extracted from the \texttt{New York Times Gigaword Corpus} using the Open Information Extraction system Ollie~\citep{schmitz2012open}. We filtered the events with frequencies less than 3 and ended up with 4,029,877 distinct events. We use the MCNC dataset adopted in \citet{lee-goldwasser-2019-multi}\footnote{\url{https://github.com/doug919/multi_relational_script_learning}} for the transfer task.

\label{sec:train_detail}
Our event representation model is implemented using the Texar-PyTorch package~\citep{hu2018texar}. The model starts from the pre-trained checkpoint of \texttt{BERT-based-uncased}~\citep{devlin-etal-2019-bert} and we use the $[\mathrm{CLS}]$ token representation as the event representation. We train our model with a batch size of 256 using an Adam optimizer. The learning rate is set as 2e-7 for the event representation model and 2e-5 for the prototype memory. We adopt the temperature $\tau=0.3$ and the numbers of prototypes used in our experiment is 10. 

\subsection{Event Similarity Tasks}
\label{sec:similarity_task}
Similarity task is a common way to measure the quality of vector representations. \citet{weber2018event} introduce two event related similarity tasks: (1) \texttt{Hard Similarity Task} and (2) \texttt{Transitive Sentence Similarity}.

\paragraph{Hard Similarity Task.}
The hard similarity task tests whether the event representation model can push away representations of dissimilar events while pulling together those of similar events. \citet{weber2018event} created a dataset (denoted as ``Original''), where each sample has two types of event pairs: one with events that should be close to each other but have very little lexical overlap, and another with events that should be farther apart but have high overlap. This dataset contains 230 event pairs. After that, \citet{ding-etal-2019-event-representation} extended this dataset to 1,000 event pairs (denoted as ``Extended''). For this task, we use Accuracy as the evaluation metric, which measures the percentage of cases where the similar pair receives a higher cosine value than the dissimilar pair.

\paragraph{Transitive Sentence Similarity.}
The transitive sentence similarity dataset~\citep{kartsaklis2014study} contains 108 pairs of transitive sentences that contain a single subject, object, and verb (e.g., \texttt{agent sell property}) and each pair in this dataset is manually annotated by a similarity score from 1 to 7. A larger score indicates that the two events are more similar. Following previous work~\citep{weber2018event,ding-etal-2019-event-representation,zheng2020incorporating}, we evaluate using the Spearman’s correlation of the cosine similarity predicted by each method and the annotated similarity score.

\subsection{Comparison methods.}
We compare our proposed approach with a variety of baselines. These methods can be categorized into three types:

\noindent
(1) \textbf{Co-occurrence}:
    \textbf{Event-comp}~\citep{weber2018event}, \textbf{Role-factor Tensor}~\citep{weber2018event} and \textbf{Predicate Tensor}~\citep{weber2018event} are models that use tensors to learn the interactions between the predicate and its arguments and are trained using co-occurring events as supervision.

\noindent
(2) \textbf{Discourse Relations}: This line of work exploits discourse relations.
    \textbf{SAM-Net}~\citep{lv2019sam} explores event segment relations, \textbf{FEEL}~\citep{lee2018feel} and \textbf{UniFA-S}~\citep{zheng2020incorporating} adopt discourse relations.
    
\noindent   
(3) \textbf{Commonsense Knowledge}:
    Several works have shown the effectiveness of using commonsense knowledge.
    \textbf{KGEB}~\citep{ding2016knowledge} incorporates knowledge graph information. \textbf{NTN-IntSent}~\citep{ding-etal-2019-event-representation} leverages external commonsense knowledge about the intent and sentiment of the event.

\paragraph{Results.}
Table~\ref{tab:similarity} reports the performance of different methods on the hard similarity tasks and the transitive sentence similarity task. The result shows that the proposed SWCC achieves the
best performance among the compared methods.  It not only
outperforms the Role-factor Tensor method that based on co-occurrence information, but also has better performance than the methods trained with additional annotations and commonsense knowledge, e.g. NTN-IntSent and UniFA-S. This implies the co-occurrence information of events is effective but underutilized by previous works, and the proposed SWCC makes better use of the co-occurrence information.

\begin{table*}[htbp]
\small
    \centering
        \begin{tabular}{llll}
        \toprule
        \multirow{2}{*}{Model} & \multicolumn{2}{c}{Hard similarity (Accuracy \%)} & Transitive sentence\\\cline{2-3}
        &  Original & Extended & similarity ($\rho$)\\
        \midrule
        SWCC& \textbf{80.9} & \textbf{72.1} & \textbf{0.82}\\
        \quad w/o Prototype-based Clustering& 77.4~(\rmark{-3.5}) & 67.4~(\rmark{-4.7}) & 0.77~(\rmark{-0.05})\\
        \quad w/o Weakly Supervised CL& 75.7~(\rmark{-5.2}) & 65.1~(\rmark{-7.0}) & 0.78~(\rmark{-0.04})\\
        \quad w/o MLM & 77.4~(\rmark{-3.5}) & 70.4~(\rmark{-1.7}) & 0.80~(\rmark{-0.02})\\
        \hdashline
        BERT~(InfoNCE) & 72.1 & 63.4 & 0.75\\
        BERT~(Margin)  & 43.5 & 51.4 & 0.67 \\
        \bottomrule
        \end{tabular}
        \caption{Ablation study for several methods evaluated on the similarity tasks.}
        \vspace{-1em}
        \label{tab:ablation}
\end{table*}
\paragraph{Ablation study.}
To investigate the effect of each component in our approach, we conduct an ablation study as reported in Table~\ref{tab:ablation}. We remove a certain component of SWCC and examine the corresponding performance of the incomplete SWCC on the similarity tasks.  
We first explore the impact of our prototype-based clustering method by removing the loss term $\mathcal{L}_{cp}$  in Eq.~\ref{eq:overal_loss}. We find that this component has a significant impact on  the transitive sentence similarity task. Removing this component causes a 0.05~(maximum)  point drop in performance on the transitive sentence similarity task. And for the weakly supervised contrastive learning method, we find that it has a strong impact on both hard similarity tasks, especially the extended hard similarity task. Removing this component causes an 7.0 point drop in performance of the model.
We also study the impact of the MLM auxiliary objective. As shown in Table~\ref{tab:ablation} the token-level MLM objective improves the performance on the extended hard similarity task modestly, it does not help much for the transitive sentence similarity task.

Next, we compare the InfoNCE against the margin loss in Table~\ref{tab:ablation}. For a fair comparison, the BERT~(InfoNCE) is trained using the InfoNCE objective only, with co-occurring events as positives and other samples in the minibatch as negatives, and the BERT~(Margin) is trained using the margin loss, with co-occurring events as positives and randomly sampled events as negatives. Obviously, BERT~(InfoNCE) achieves much competitive results on all tasks, suggesting that the InfoNCE with adjustable temperature works better than the margin loss. This can be explained by the fact that the InfoNCE weighs multiple different negatives, and an appropriate temperature can help the model learn from hard negatives, while the margin loss uses only one negative and can not weigh the negatives by their relative hardness.

\subsection{Transfer Task}
\label{sec:transfer_task}
We test the generalization of the event representations by transferring to a downstream event related tasks, the Multiple Choice Narrative Cloze~(MCNC) task~\citep{granroth2016happens}, which was proposed to evaluate script knowledge. In particular, given an event chain which is a series of events, this task requires a reasoning system to distinguish the next event from a small set of randomly drawn events. 
We evaluate our methods with several methods based on unsupervised learning: (1) \textbf{Random} picks a candidate at random uniformly; (2) \textbf{PPMI}~\citep{chambers-jurafsky-2008-unsupervised} uses co-occurrence information and calculates Positive PMI for event pairs; (3) \textbf{BiGram}~\citep{jans2012skip} calculates bi-gram conditional probabilities based on event term frequencies; (4) \textbf{Word2Vec}~\citep{mikolov2013efficient} uses the word embeddings trained by Skipgram algorithm and event representations are the summation of word embeddings of predicates and arguments.
Note that we did not compare with supervised methods~\cite{bai-etal-2021-integrating,zhou-etal-2021-modeling,lv-etal-2020-integrating} since unsupervised ones are more suitable for purely evaluating event representations.

\paragraph{Results.}
Table~\ref{tab:mcnc} reports the performance of different methods on the MCNC task. As shown in the table, SWCC achieves the best accuracy on the MCNC task under the zero-shot transfer setting, suggesting the proposed SWCC  has better generalizability to the downstream tasks than other compared methods. 

\begin{table}[htbp]
\small
\centering
    \begin{tabular}{lr}
    \toprule
    Model &  Accuracy (\%) \\
    \midrule
    Random & 20.00 \\
    PPMI* & 30.52 \\
    BiGram* & 29.67 \\
    Word2Vec* & 37.39 \\
    \hdashline
    BERT~(Margin) & 36.50\\
    BERT~(InfoNCE) & 39.23 \\
    SWCC& \textbf{44.50} \\
    \bottomrule
    \end{tabular}
    \caption{Evaluation performance on the MCNC task. Best results are bold. *: results reported in the previous work~\citep{lee-goldwasser-2019-multi}.}
    \vspace{-1.5em}
    \label{tab:mcnc}
\end{table}



\section{Analysis and Visualization}
\begin{figure}[htbp] 
  \centering
  \includegraphics[width=0.9\linewidth]{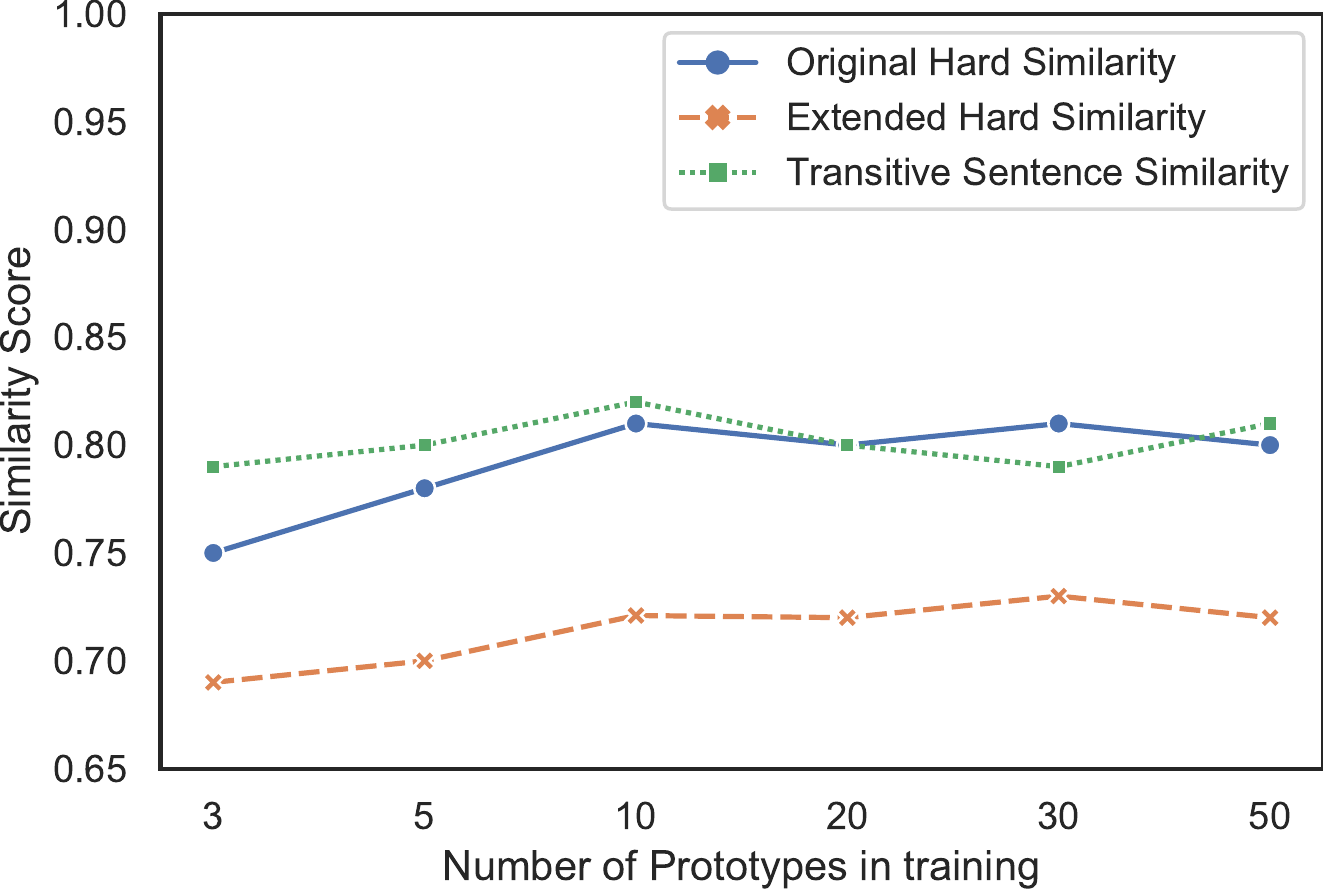}
  \caption{Impact of \# of Prototypes}
  \vspace{-1em}
  \label{fig:cluster}
\end{figure}

\begin{figure}[htbp]
  \centering
  \subfigure{
  \begin{minipage}[t]{0.5\linewidth}
  \centering
  \includegraphics[scale=0.32]{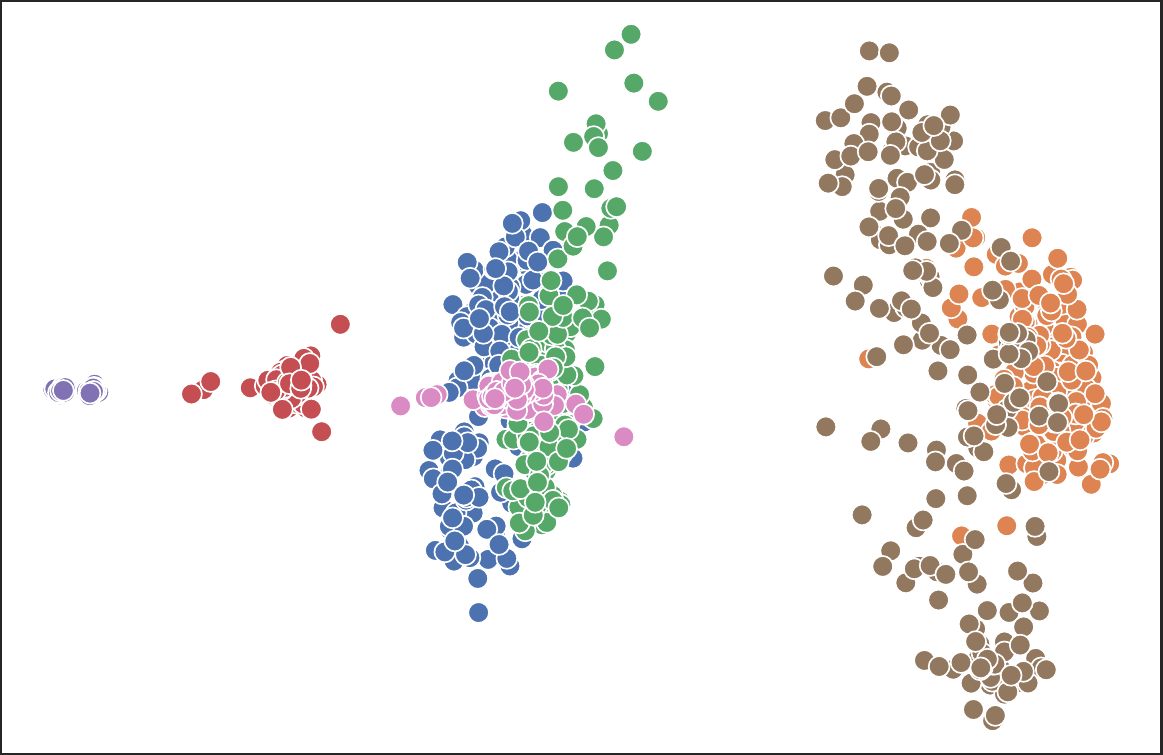}
  \end{minipage}%
  }%
  \subfigure{
  \begin{minipage}[t]{0.5\linewidth}
  \centering
  \includegraphics[scale=0.32]{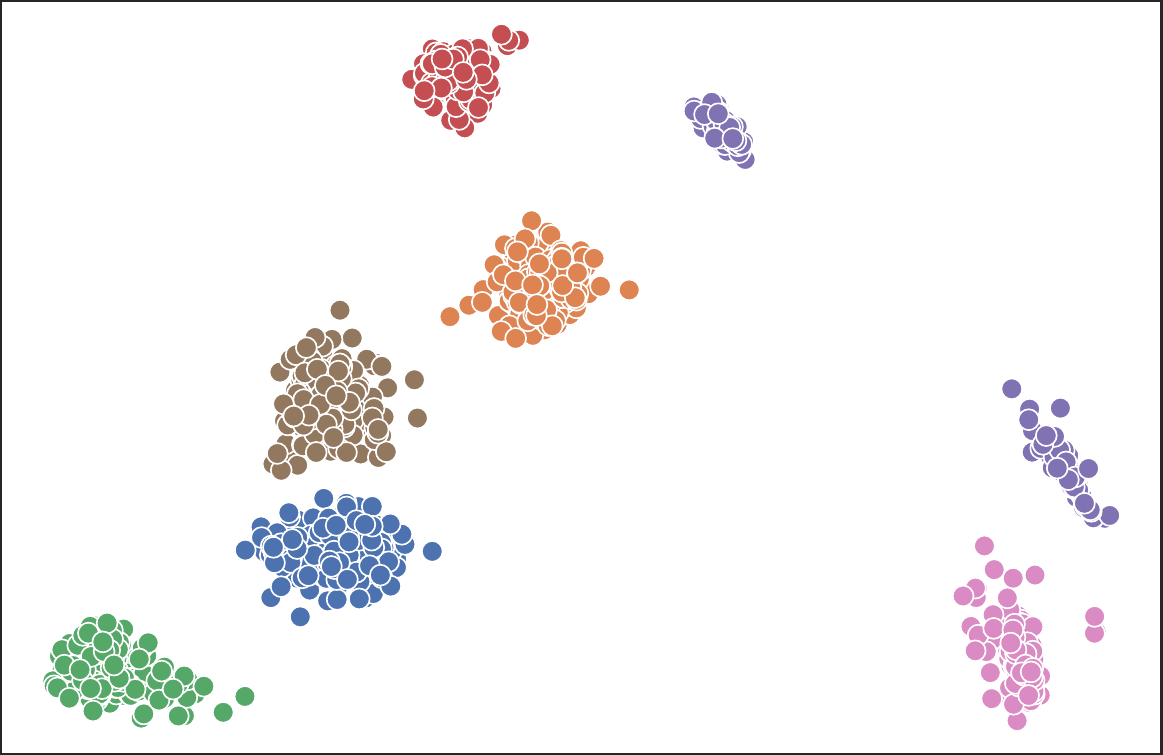}
  \end{minipage}%
  
  }%
  \caption{2D visualizations of the event representation spaces learned by BERT~(InfoNCE)~(left) and SWCC~(right), respectively. Each event is denoted by a color indicating a prototype.}
  \label{fig:pca_vis}
  \vspace{-1em}
\end{figure}

In this section, we further analyze the prototype-based clustering method.
\paragraph{Number of prototypes.}
Figure~\ref{fig:cluster} displays the impact of the number of prototypes in training. As shown in the figure, the performance increases as the number $M$ increases, but it will not further increase after 10. We speculate that because these evaluation data are too small and contain too few types of relations, a larger number of prototypes would not help much in performance improvement.
\paragraph{Visualization of learned representation.}
We randomly sample 3000 events and embed the event representations learned by BERT~(InfoNCE) and SWCC in 2D using the PCA method. The cluster label of each event is determined by matching its representation to the set of $M$ prototypes.
The resulting visualizations are given in Figure~\ref{fig:pca_vis}. It shows that the proposed SWCC yields significantly better clustering performance than the BERT~(InfoNCE), which means, to a certain extent, the prototype-based clustering method can help the event representation model capture various relations of events.
Overall, the class separation in the visualizations qualitatively agrees with the performance in Table~\ref{tab:similarity}.

\paragraph{Case study.} We also present sampled events from two different prototypes in Table~\ref{tab:case} (see Appendix for more examples), to further demonstrate the ability of SWCC to capture various relations of events. We can see that the events belonging to ``Prototype1'' mainly describe financial stuff, for example, ``earnings be reduced'', while the events belonging to ``Prototype2'' are mainly related to politics. Clearly, the events in the same cluster have the same topic. And we also find that there are also causal and temporal relations between some of these events. For example, ``earnings be reduced'' led to ``company cut costs''.
\begin{table}[htbp]
    \small
    \centering
    \begin{tabular}{ll}
        \toprule
        \makecell[c]{\textbf{Prototype1}}&\makecell[c]{\textbf{Prototype2}}\\
        \midrule
        loans be sell in market&president asked senate\\
        earnings be reduced & he deal with congress\\
        company cut costs &senate reject it\\
        earnings be flat&council gave approval\\
        banks earn fees&council rejected bill\\
        \bottomrule
    \end{tabular}
    \caption{Example events of two different prototypes.}
    \vspace{-1.5em}
    \label{tab:case}
\end{table}

\section{Related Work}
\paragraph{Event representation learning.}
Effectively representing events and their relations~(casual, temporal, entailment~\cite{ning-etal-2018-multi,yu2020enriching}) becomes important for various downstream tasks, such as event schema induction~\cite{li2020connecting}, event narrative modeling~\cite{chambers-jurafsky-2008-unsupervised,li2018constructing,lee-goldwasser-2019-multi}, event knowledge graph construction~\cite{sap2019atomic,zhang2020aser} etc.
Many efforts have been devoted into learning distributed event representation. Though driven by various motivations, the main idea of these methods is to exploit explicit relations of events as supervision signals and these supervision signals can be roughly categorized into three types:
(1) discourse relations (e.g. casual and temporal relations) obtained with automatic annotation tools~\citep{zheng2020incorporating};
(2) manually annotated external knowledge (e.g. sentiments and intents) ~\citep{lee2018feel,ding-etal-2019-event-representation} and 
(3) co-occurrence information~\citep{weber2018event}.
Existing work has focused on the first two supervision signals, with less research on how to better utilize co-occurrence information. 
Though, discourse relations and external knowledge are fine-grained relations that can provide more accurate knowledge, the current explicitly defined fine-grained relations fall under a small set of event relations. Co-occurrence information is easily accessible but underutilized. Our work focus on exploiting document-level co-occurrence information of events to learn event representations, without any additional annotations.

\paragraph{Instance-wise contrastive learning.}
Recently, a number of instance-wise contrastive learning methods have emerged to greatly improve the performance of unsupervised visual and text representations~\citep{he2020momentum,chen2020improved,chen2020simple,chen2021exploring,grill2020bootstrap,zbontar2021barlow,chen2020simple,hu2021adco,gao2021simcse,yang-etal-2021-contrastive-representation}. This line of work aims at learning an embedding space where samples from the same
instance are pulled closer and samples from different instances are pushed apart, and  usually adopt InfoNCE~\citep{oord2019representation} objective for training their models. Unlike the margin loss using one positive example and one negative example, the InfoNCE can handle the case where there exists multiple negative samples. In our work, we extend the InfoNCE, which is a self-supervised contrastive learning approach, to a weakly supervised contrastive learning setting, allowing us to effectively leverage co-occurrence information. 

\paragraph{Deep unsupervised clustering.}
Clustering based methods have been proposed for representation learning~\citep{caron2018deep,zhan2020online,caron2020unsupervised,li2020prototypical,zhang2021supporting}. \citet{caron2018deep} use k-means assignments
pseudo-labels to learn visual representations. Later,  
\citet{asano2019self} and \citet{caron2020unsupervised} cast the pseudo-label assignment problem as an instance of the
optimal transport problem. Inspired by \citet{caron2020unsupervised}, we leverage a similar formulation to map event representations to prototype vectors. Different from \citet{caron2020unsupervised}, we
simultaneously perform weakly supervised contrastive learning and prototype-based clustering.

\section{Conclusion}
In this work, we propose a simple and effective framework~(\textbf{SWCC}) that learns event representations by making better use of co-occurrence information of events, without any addition annotations. 
In particular, we introduce a weakly supervised contrastive learning method that allows us to consider multiple positives and multiple negatives, and a prototype-based clustering method that avoids semantically related events being pulled apart.
Our experiments indicate that our approach not only outperforms other baselines on several event related tasks, but has a good clustering performance on events. We also provide a thorough analysis of the prototype-based clustering method to demonstrate that the learned prototype vectors are able to implicitly capture various relations between events.

\section*{Acknowledgements}
This work was partially supported by the National Natural Science Foundation of China (61876053, 62006062, 62176076), the Shenzhen Foundational Research Funding (JCYJ20200109113441941, JCYJ20210324115614039), Joint Lab of Lab of HITSZ and China Merchants Securities.

\bibliography{custom}
\bibliographystyle{acl_natbib}

\clearpage
\appendix
\section{Appendix}

\subsection{Model Analysis}

\paragraph{Impact of Temperature.}
We study the impact of the temperature by trying out different temperature rates in Table~\ref{tab:temperature} and observe that all the variants underperform the $\tau=0.3$.
\begin{table}[htbp]
\scriptsize
    \centering
        \begin{tabular}{lrrr}
        \toprule
        \multirow{2}{*}{SWCC} & \multicolumn{2}{c}{Hard similarity (Acc. \%)} & Transitive sentence\\\cline{2-3}
        &  Original & Extended & similarity ($\rho$)\\
        \midrule
        with Temperature &&&\\
        \quad $\tau=0.2$ & 80.0 & 71.0 & 0.80\\
        \quad $\tau=0.3$ & \textbf{80.9} & \textbf{71.3} & \textbf{0.82}\\
        \quad $\tau=0.5$ & 77.4 & 68.7 & 0.78\\
        \quad $\tau=0.7$ & 72.2 & 50.5 & 0.75\\
        \quad $\tau=1.0$ & 48.7 & 22.9 & 0.67\\
        \bottomrule
        \end{tabular}
        \caption{Impact of Temperature~($\tau$).}
        \label{tab:temperature}
\end{table}
\paragraph{Impact of the MLM objective with different $\gamma$.}
Table~\ref{tab:gamma} presents the results obtained with different $\gamma$. As can be seen in the table, larger or smaller values of gamma can harm the performance of the model. $\gamma=1.0$ gives a better overall performance of the model.
\begin{table}[htbp]
\scriptsize
    \centering
        \begin{tabular}{lrrr}
        \toprule
        \multirow{2}{*}{SWCC} & \multicolumn{2}{c}{Hard similarity (Acc. \%)} & Transitive sentence\\\cline{2-3}
        &  Original & Extended & similarity ($\rho$)\\
        \midrule
        with MLM &&& \\
        \quad$\gamma=0.1$ & 76.5 & 70.9 & 0.80\\
        \quad$\gamma=0.5$ & 79.1 & 71.1 & 0.81\\
        \quad$\gamma=1.0$ & \textbf{80.9} & \textbf{72.1} & \textbf{0.82}\\
        \quad$\gamma=1.5$ & \textbf{80.9} & 71.9 & 0.81\\
        \quad$\gamma=2.0$ & \textbf{80.9} & \textbf{72.1} & 0.80\\
        \bottomrule
        \end{tabular}
        \caption{Impact of the MLM objective with different $\gamma$.}
        \label{tab:gamma}
\end{table}

\paragraph{Impact of the prototype-based clustering objective with different $\beta$.}
Finally, we study the impact of the prototype-based clustering objective with different $\beta$. As can be seen in the Table~\ref{tab:beta}, the larger the $beta$, the better the performance of the model on the hard similarity task.
\begin{table}[htbp]
    \scriptsize
    \centering
        \begin{tabular}{lrrr}
        \toprule
        \multirow{2}{*}{SWCC} & \multicolumn{2}{c}{Hard similarity (Acc. \%)} & Transitive sentence\\\cline{2-3}
        &  Original & Extended & similarity ($\rho$)\\
        \midrule
        with $\mathcal{L}_{pc}$ &&& \\
        \quad $\beta=0.01$ & 78.3 & 71.6 & 0.80\\
        \quad$\beta=0.05$ & 76.5 & 71.6 & 0.80\\
        \quad$\beta=0.1$ & \textbf{80.9} & 72.1 & \textbf{0.82}\\
        \quad$\beta=0.3$ & \textbf{80.9} & 71.3 & \textbf{0.82}\\
        \quad$\beta=0.5$ & \textbf{80.9} &\textbf{ 73.1} & 0.80\\
        \quad$\beta=0.7$ & \textbf{80.9} & 72.8 & 0.80\\
        \quad$\beta=1.0$ & \textbf{80.9} & 72.1 & 0.80\\
        \bottomrule
        \end{tabular}
        \caption{Impact of the prototype-based clustering objective with different $\beta$.}
        \label{tab:beta}
\end{table}

\subsection{Case Study}
\paragraph{Case study.} We present sampled events from six different prototypes in Table~\ref{tab:full_case} to further demonstrate the ability of SWCC to capture various relations of events. We can see that the events belonging to ``Prototype1'' mainly describe financial stuff, for example, ``earnings be reduced'', while the events belonging to ``Prototype2'' are mainly related to politics. Clearly, the events in the same cluster have the same topic. And we also find that there are also causal and temporal relations between some of these events. For example, ``earnings be reduced'' leads to ``company cut costs''.
\begin{table*}[htbp]
    \small
    \centering
    \begin{tabular}{lll}
        \toprule
        \makecell[c]{\textbf{Prototype1}}&\makecell[c]{\textbf{Prototype2}}&\makecell[c]{\textbf{Prototype3}}\\
        \midrule
        loans be sell in market&president asked senate&he be known as director\\
        earnings be reduced & he deal with congress&Wright be president of NBC\\
        company cut costs &senate reject it&Cook be chairman of ARCO\\
        earnings be flat&council gave approval&Bernardo be manager for Chamber\\
        banks earn fees&council rejected bill&Philbin be manager of Board\\
        \midrule
        \makecell[c]{\textbf{Prototype4}}&\makecell[c]{\textbf{Prototype5}}&\makecell[c]{\textbf{Prototype6}}\\
        \midrule
        he be encouraged by things&kind is essential&Dorsey said to James\\
        I be content&it be approach to life&Gephardt said to Richard\\
        they be motivated by part&we respect desire&Pherson said to Kathy\\
        they be meaningful&thing be do for ourselves&Stone said to Professor\\
        he be ideal&it be goal of people&Stiles said to Thomas\\
        \bottomrule
    \end{tabular}
    \caption{Example events of different prototypes.}
    \vspace{-1.5em}
    \label{tab:full_case}
\end{table*}

\end{document}